\def\year{2021}\relax
\documentclass[letterpaper]{article} 
\usepackage{aaai21}  
\usepackage{times}  
\usepackage{helvet} 
\usepackage{courier}  
\usepackage[hyphens]{url}  
\usepackage{graphicx} 
\usepackage{natbib}  
\usepackage{caption} 
\usepackage{hyperref}
\title{Shallow-UWnet : Compressed models for underwater image enhancement}
\author{
    Ankita Naik,\textsuperscript{\rm 1}\thanks{equal contribution}
    Apurva Swarnakar,\textsuperscript{\rm 1*} 
    Kartik Mittal\textsuperscript{\rm 1*}
    \\
}
\affiliations{
    \textsuperscript{\rm 1}University of Massachusetts Amherst\\

    Amherst, MA-01002\\
    \{arnaik, aswarnakar, kartikmittal\}@umass.edu\\

}

\begin{document}

\maketitle

\begin{abstract}
Over the past few decades, underwater image enhancement has attracted increasing amount of research effort due to its significance in underwater robotics and ocean engineering. Research has evolved from implementing physics-based solutions to using very deep CNNs and GANs. However, these state-of-art algorithms are computationally expensive and memory intensive. This hinders their deployment on portable devices for underwater exploration tasks. These models are trained on either synthetic or limited real world datasets making them less practical in real-world scenarios. In this paper we propose a shallow neural network architecture, \textbf{Shallow-UWnet} which maintains performance and has fewer parameters than the state-of-art models. We also demonstrated the generalization of our model by benchmarking its performance on combination of synthetic and real-world datasets.The dataset and code are available at \href{https://github.com/mkartik/Shallow-UWnet}{https://github.com/mkartik/Shallow-UWnet}.
\end{abstract}

\section{Introduction}
Underwater robotics represent a fast-growing research area. Recently, great efforts are being made in developing autonomous underwater vehicles deployed (AUVs) to tackle challenging engineering  problems such as underwater surveillance, seabed mapping, underwater archaeological exploration, garbage collection, underwater rescue operations and military operations. Many of these applications require real-time interpretation of images / videos for the AUV to intelligently perceive the environment and take follow-up measures.\\
\\
Inherently, underwater images are degraded by wavelength dependent absorption and forward and backward scattering due to particles in the water. These degraded images have lower visibility, reduced contrast, color deviations and even introduce color casts, which limit their applicability in underwater vision based downstream tasks like tracking, classification and detection. In addition, underwater images have a dominating green or blue hue as the red wavelengths tend to get absorbed in deep water.\\
\\
To handle the above mentioned issues, first step before any downstream underwater image interpretation tasks is image improvement which encompasses tasks of image enhancement, image restoration and supplementary task specific methods. In our study we are focusing on the image enhancement task which helps in alleviating the problems by restoring the perceptual and statistical qualities of the image in real-time.\\
\\
Underwater image enhancement methods extract image information without any prior knowledge about the environment making them more generalized than image restoration methods. The existing literature based on very deep Convolutional Neural Networks (CNNs) and Generative Adversarial Network (GAN)-based models focus on aspects such as as noise removal, contrast stretch, combined improvement with multi-information and deep learning for image dehazing. However, the high computational and memory requirements of these big models makes them heavy for the real-time underwater image enhancement tasks. Thus, to improve the deployability of machine learning models by reducing the compute and memory requirement, meanwhile maintaining comparable performance with state of the art models, we propose \textbf{Shallow-UWnet}.

\section{Background}
In the past few years variety of methods have been proposed for image enhancement task which can be broadly classified into three groups: Non-physical model, physical model-based method and deep learning methods.\\
\\
While Non-physical models improve by adjusting image pixel values rather than a mathematical equation, Physical models formulates the degradation process of the image by estimating the parameters of the model. But both are not sufficient for underwater image enhancement as they ignore specific underwater properties.\\
\\
Deep Learning Methods tend to perform better as they solely focus on color correction. These majorly revolve around generative adversarial networks, convolutional neural networks. We have reviewed current state-of-art GAN based model and CNN-based models in detail.\\
\\
\textbf{FUnIE-GAN} \cite{islam2020simultaneous} : It deblurs the images by formulating to an image-to-image translation problem, assuming that there exists a non-linear mapping between the distorted and enhanced images. Further, a conditional GAN-based model learns this mapping by adversarial training on a large-scale dataset. But it incorrectly models sunlight and amplifies the noise in the background and leads to either over-saturated or under-saturated images.\\
\\
\textbf{Water-Net} \cite{li2019underwater} : It is a gated fusion CNN trained by on UIEB \cite{Li_2017} for underwater image enhancement. To align with the characteristics of the degraded underwater images, there are three enhanced inputs to the Water-Net model generated by applying White Balance (WB), Gamma Correction(GC) and Histogram Equalizaion (HE). But it is a complex CNN architecture, which suffers from the effect of backscatter.\\
\\
\textbf{UResnet} \cite{liu2019underwater} : It is a CNN based residual network which is a more comprehensive supervised learning method for underwater image enhancement.

\section{Underwater Datasets}
We explored a variety of underwater image datasets ranging from dataset synthetically hazzed to real-world underwater images. We utilized three of them detailed below to benchmark the generalization of our model against state-of-art models trained on these.\\
\begin{enumerate}
    \item \textbf{EUVP Dataset} \cite{islam2020fast}\\
    The EUVP Dataset (Enhancement  of  Underwater  Visual  Perception)  is a large collection of 10K paired and 25K unpaired images of poor and good perceptual quality, captured by the authors \cite{islam2020fast} during oceanic explorations under various visibility conditions. The paired images were generated by distorting real world images using a underwater distortion model based on CycleGAN. We use the ImageNet paired images, part of the EUVP dataset for our model training-validation and another set of EUVP-Dark which are a pair of highly hazzed images for our model testing.\\
    \item \textbf{UIEB Dataset} \cite{li2019underwater}\\ The UIEB dataset  (Underwater Image Enhancement Benchmark) consists of 890 real underwater images that were captured under different lighting conditions and have diverse color range and degrees of contrast. The authors have provided respective reference images that are color cast-free (have relatively genuine color) and have improved visibility and brightness compared to the source image. We test our model on this dataset, as this dataset acts as a real world underwater dataset with reference images, obtained without synthetic techniques.\\
    
    \item \textbf{UFO-120 Dataset} \cite{islam2020simultaneous}\\
    The UFO-120 dataset comprises of 1620 paired underwater images that were collected during oceanic explorations in different water types. Style-transfer techniques were then used by the authors to generate the respective distorted images. The dataset can act as an evaluation for both underwater image enhancement as well as underwater image super resolution models. We test our model on the 120 paired underwater images labelled as the testing set in the UFO-120 dataset.
\end{enumerate}

\section{Evaluation Metrics}
We quantitatively evaluate the output images of our model using the standard metrics \cite{yang2019depth} namely Peak Signal-to-Noise Ratio (PSNR) and Structural Similarity Index Measure (SSIM). The SSIM and PSNR quantify the structural similarity and reconstruction quality of the enhanced output image with the respective reference image. We also analyse the quality of the generated output images using a non-reference underwater image quality measure (UIQM). The UIQM comprises of three underwater image attributed measures: image colorfulness (UICM), sharpness (UISM), and contrast (UIConM), wherein each attribute evaluates one aspect of underwater image degradation. The UIQM metrics is given by \\
\begin{equation}
    \mbox{UIQM} = c_1 \times \mbox{UICM} + c_2 \times \ \mbox{UISM} 
+ c_3 \times \mbox{UIConM}
\end{equation}
with parameters $c_1=0.0282, c_2=0.2953, c3 = 3.5753$ set according to \cite{panetta2015human}  paper.\\

We measured the quality of model compression and acceleration, using compression and the speedup rates \cite{cheng2017survey}. \\
\\
$$\mbox{Compression rate }(M, M_*) = \frac{\alpha(M)}{\alpha(M_*)}$$
\\
$$\mbox{Speed-up rate }(M, M_*) = \frac{\beta(M)}{\beta(M_*)}$$
where, \\
 $\alpha(M) $ are the number of parameters of the model M, \\
 $\beta(M) $ is testing time for one image for the model M$, \\
 M$ is the original model,\\
 $M_*$ is the compressed model

\section{Our Proposed Approach}
In this section, we first discuss the details of the proposed architecture, the network calculation and then the evaluation metrics. 
\begin{figure}[h]
    \centering
    \includegraphics[width=.9\linewidth]{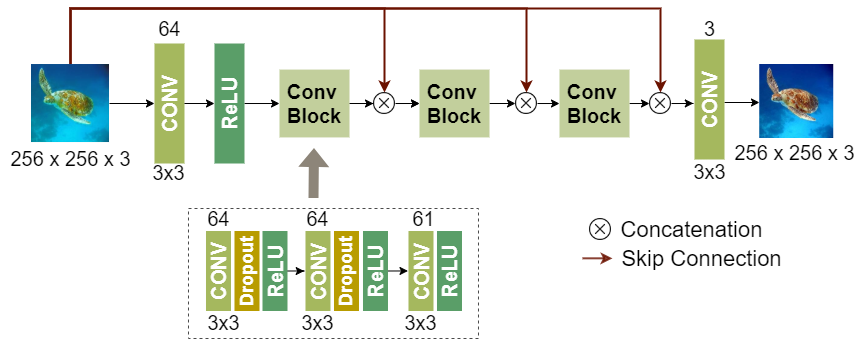}
    \caption{Model Architecture}
    \label{fig:arch}
\end{figure}

\subsection{Network Architecture}
Figure  \ref{fig:arch}  shows  the  architecture  diagram  of  our  proposed Shallow-UWnet model. The model is comprised of a fully connected convolution network connected to three densely connected convolutional blocks in series. The input image is being concatenated to the output of each block using a skip connection. The input to our model is a $256 \times 256$ RGB underwater image.
The raw input image is passed through the first layer of convolution layer with kernel size 3x3 to generate 64 feature maps, followed by a ReLU activation layer, then chained with three convolution blocks. A final convolution layer with 3 kernels generates the enhanced underwater image. 

\subsubsection{Convolutional Blocks}The ConvBlocks consists of two sets of convolution layers, each followed by a dropout and ReLU activation function. The output is then passed through another set of Conv-Relu pair which facilitates concatenation of the raw image from the skip connection. The series of ConvBlocks along with skip connection act as a deterrent for overfitting of the network over the train data. Thus, supporting generalization of the network.

\subsubsection{Skip Connections: }The raw input image is concatenated to the output of each residual block via skip connections. In the scenario of encountering vanishing gradient issue such a skip connection would impose larger weight on the channels associated with raw input image as compared to the channels output from the ConvBlock. Thereby, ensuring feature learning from each block along with 
incorporating essential characteristics from the base raw image.

\subsection{Network Loss}
The model is trained using a multi-term loss function to preserve sharpness of the edges, impose structural and texture similarity of the enhanced image, and to account for the pixel-wise loss generated. It is calculated using the following two loss components:
\begin{enumerate}
    \item \textit{MSE Loss: }  The pixel-wise mean squared error (MSE) loss computes the sum of squared differences between the the estimated image I, and the clear ground truth image, I*.
        $$L_{MSE} = \frac{1}{N}\sum_{i=1}^{N} (I_i - I^*_i)^2$$
    \item \textit{VGG Perpetual Loss: } The perceptual loss \cite{johnson2016perceptual}  is defined based on the ReLU activation of the pretrained 19 layers VGG Network. The enhanced image and the ground truth image are passed to the last convolutional layer of the pretrained VGG network to get the feature representations. The perceptual loss is calculated as the distance between the feature representations of the enhanced image, I, and the ground truth clear image, I*, which is denoted by $L_{VGG}$.\\ 
    \item The final loss L, is calculated as the summation of the two losses.
    	 $$L_{TOTAL} = L_{MSE}  + L_{VGG}$$
\end{enumerate}

\section{Experimental Evaluations}
To evaluate our model, we performed qualitative and quantitative comparisons with the recent state-of-the-art underwater image enhancement methods on synthetic and real-world underwater images. These methods include WaterNet \cite{li2019underwater}, FunIE-GAN \cite{islam2020fast} and Deep SESR \cite{islam2020simultaneous}. We used model checkpoints provided by the corresponding authors to produce the best results for an objective evaluation.\\

\begin{enumerate}
\item \textbf{Training and Validation Data}\\
EUVP Underwater ImageNet dataset was used for training and validation purposes. This dataset images were collected using a variety of cameras such as GoPros, low light USB etc during oceanic explorations under various visibility conditions and the corresponding paired images are generated using CycleGAN by \cite{islam2020fast}. Since EUVP dataset captures locations and perceptual quality diversity, we chose this dataset so that our model can be generalized to other underwater datasets. Our model is trained on 6128 images while 515 are used for validation. The input images are of various resolutions $800 \times 600$, $640 \times 480$, $256 \times 256$, and $224 \times 224$ which are resized to $256 \times 256$ before training the model.\\

\item \textbf{Network Implementation and Training}\\
Our model was trained using ADAM optimizer with learning rate set to 0.0002 and layers dropout to 0.2. The batch size is set to 1. It takes around ten hours to optimize a model over 50 epochs.
We used PyTorch as the deep learning framework on an Intel(R) Core(TM) i7-8750H CPU, 16GB RAM, and a Nvidia GTX 1060 GPU\\

\item \textbf{Testing Data Sets}
We tested our model on variety of synthetic and real world images to benchmark the transferability of our model to different datasets.
\begin{enumerate}
\item UFO-120 \cite{islam2020simultaneous}: Clear images were collected from oceanic explorations in different water types and style transfer was used to generate the respective distorted image. A subset of 120 such images are used for testing

\item EUVP Dark \cite{islam2020fast}: EUVP has clustered 5500 paired images with dark underwater background under a separate dataset. We used a subset of 1000 such images for testing our model

\item UIEB \cite{li2019underwater}:  We used Underwater Image Enhancement Benchmark Dataset (UIEBD) as to simulate the real-world underwater scenes. It consists of 890 paired underwater images which were captured under different light conditions and have diverse color range and degrees of contrast. The reference images were generated using meticulous pairwise comparisons.
\end{enumerate}
\end{enumerate}

\begin{table*}[t]
    \centering
    \begin{tabular}{| c || c | c | c | c | c |}
    \hline
\textbf{Metric} & \textbf{Datasets} & WaterNet & FUnIE-GAN & Deep SESR & Shallow-UWnet \\
    \hline
  & EUVP-Dark & $24.43 \pm 4.64$ & $26.19 \pm 2.87$ & $25.30 \pm 2.63$ & $\textbf{27.39} \pm \textbf{2.70}$\\
    \cline{2-6}
PSNR & UFO-120 & $23.12 \pm 3.31$ &$24.72 \pm 2.57$ & $\textbf{26.46} \pm \textbf{3.13}$ & $25.20 \pm 2.88$\\
 \cline{2-6}
 & UIEB & $19.11 \pm 3.68$ &$19.13 \pm 3.91$ & $\textbf{19.26} \pm \textbf{3.56}$ & $18.99 \pm 3.60$\\

    \hline
  & EUVP-Dark & $0.82 \pm 0.08$ & $0.82 \pm 0.08$ & $0.81 \pm 0.07$ & $\textbf{0.83} \pm \textbf{0.07}$\\
    \cline{2-6}
SSIM & UFO-120 & $0.73 \pm 0.07$ & $0.74 \pm 0.06$ & $\textbf{0.78} \pm \textbf{0.07}$ & $0.73 \pm 0.07$\\
 \cline{2-6}
 & UIEB & $\textbf{0.79} \pm \textbf{0.09}$ & $0.73 \pm 0.11$ & $0.73 \pm 0.11$ & $0.67 \pm 0.13$\\
 
    \hline
  & EUVP-Dark & $2.97 \pm 0.32$ & $2.84 \pm 0.46$ & $2.95 \pm 0.32$ & $\textbf{2.98} \pm \textbf{0.38}$\\
    \cline{2-6}
UIQM & UFO-120 & $2.94 \pm 0.38$ & $2.88 \pm 0.41$ & $\textbf{2.98} \pm \textbf{0.37}$ & $2.85 \pm 0.37$\\
 \cline{2-6}
 & UIEB & $\textbf{3.02} \pm \textbf{0.34}$ & $2.99 \pm 0.39$ & $2.95 \pm 0.39$ & $2.77 \pm 0.43$\\
     \hline
    \end{tabular}
    \caption{Underwater Image Enhancement Performance Metric}
    \label{tab:enhancement}
\end{table*}

\begin{table*}[t]
    \centering
    \begin{tabular}{| c || c | c | c | c |}
    \hline
 \textbf{Models} & \# Parameters & Compression Ratio & Testing per image (secs) & Speed-Up\\
    \hline
    Our Model & 2,19,840 & 1 & 0.02 & 1 \\
    \cline{1-5}
    WaterNet & 10,90,668 & 3.96 & 0.50 & 24 \\
    \cline{1-5}
    Deep SESR & 24,54,023 & 10.17 & 0.16 & 7 \\ 
    \cline{1-5}
    FUnIE-GAN & 42,12,707 & 18.17 & 0.18 & 8\\
     \hline
    \end{tabular}
    \caption{Model Compression Performance metric as mentioned in \cite{cheng2017survey}}
    \label{tab:compression}
\end{table*}

\section{Results}

\begin{figure*}[t]
    \centering
    \includegraphics[width=.9\linewidth]{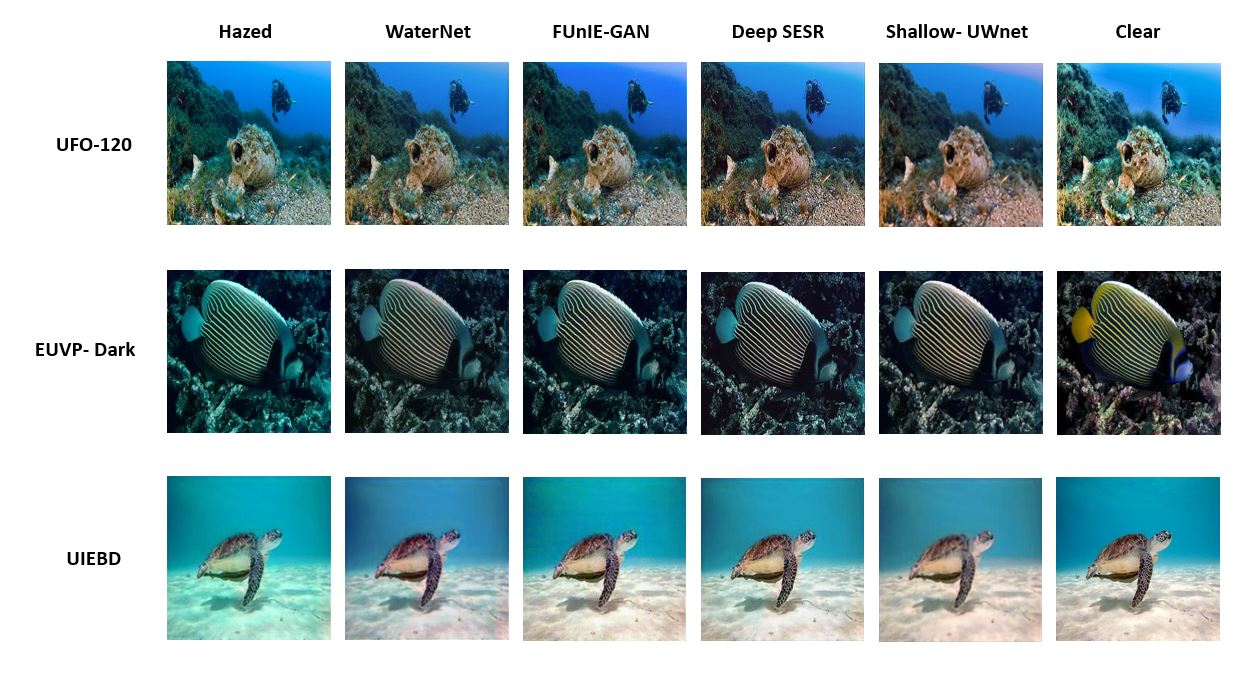}
    \caption{Underwater Image Enhancement for multiple datasets and models}
    \label{fig:examples}
\end{figure*}

We analyse the experimental results both quantitatively and qualitatively in this section.

Our  proposed Shallow-UWnet  model is able to achieve comparable performance across all three test datasets as in Table \ref{tab:enhancement}.\\
When analyzed for the model compression and acceleration, Shallow-UWnet has less number of trainable parameters than all three state of art models making it lighter for the on device deployment across different locations and conditions, Table \ref{tab:compression}.\\
It is also able to process the test images much faster making it feasible for the real time underwater image enhancement applications.

Further analyzing the results for each test dataset, Table \ref{fig:arch}, following observations are made:
\begin{enumerate}
    \item \textbf{EUVP- Dark:}\\
    Shallow-UWnet outperforms on the PSNR, SSIM and UIQM metrics as compared to all other models.Even though Shallow-UWnet was trained on the images with better lighting conditions, it is able to clean the color hue and sharpen the images. This emphasizes on the generalizing nature of the Shallow-UWnet.
    \item \textbf{UFO-120}:\\
    Since Deep SESR was trained on UFO-120, it is has the best performance of all the models on the same. Even though the UFO dataset creation ranks very different than EUVP-Imagenet dataset (data used to train Shallow-UWnet) we see that Shallow-UWnet ranks second best proving its capabilities on synthetic data.
    \item \textbf{UIEB Dataset}:\\
    Shallow-UWnet was also able to similarly performace well on this near real-world data. Since WaterNet was trained on the UIEBD,it gives better performance than Shallow-UWnet on the same. Despite being trained on synthetic datasets, Shallow-UWnet and Deep SESR performances lie in similar ranges for this dataset.
\end{enumerate}

A comparative visual analysis of the performance of these models on the three datasets can be seen in Figure \ref{fig:examples}

\section{Conclusion}
The use of convolutional neural networks has widespread applications in computer vision and this understanding can also be extended to underwater images. Our  proposed Shallow-UWnet  model maintains comparable quantitative performance while requiring 18 times lesser trainable parameters and makes testing 10 times faster. It is noteworthy that our model is able to generalize on varied datasets emphasizing its real world application.

{\small
\bibliography{bibtex}
}

\end{document}